%% file: main.tex
\documentclass{article} 
\usepackage[final]{eiml_style_2025}
\usepackage{natbib}
\usepackage[]{hyperref}
\usepackage{amssymb}
\usepackage{amsmath}
\usepackage{mathtools}
\usepackage{url}
\usepackage{tikz}
\usetikzlibrary{positioning, arrows.meta, automata}

\title{On the Notion that Language Models Reason}

\author{
Bertram Højer \\
Department of Computer Science\\
IT University of Copenhagen\\
\texttt{berh@itu.dk} \\
}

\begin{document}

\maketitle

\begin{abstract}
Language models (LMs) are said to be exhibiting reasoning, but what does this entail? We assess definitions of \textit{reasoning} and how key papers in the field of natural language processing (NLP) use the notion and argue that the definitions provided are not consistent with how LMs are trained, process information, and generate new tokens. To illustrate this incommensurability we assume the view that transformer-based LMs implement an \textit{implicit} finite-order Markov kernel mapping contexts to conditional token distributions. In this view, reasoning-like outputs correspond to statistical regularities and approximate statistical invariances in the learned kernel rather than the implementation of explicit logical mechanisms. This view is illustrative of the claim that LMs are ``statistical pattern matchers'' and not \textit{genuine reasoners} and provides a perspective that clarifies why reasoning-like outputs arise in LMs without any guarantees of logical consistency. This distinction is fundamental to how epistemic uncertainty is evaluated in LMs. We invite a discussion on the importance of how the computational processes of the systems we build and analyze in NLP research are described.
\end{abstract}

\input{tex/01-introduction}
\input{tex/02-reasoning}
\input{tex/03-markov}
\input{tex/04-discussion}
\input{tex/05-conclusion}

\bibliographystyle{plainnat}
\bibliography{references}

\end{document}

%% file: tex/01-introduction.tex
\section{Introduction}

Language models (LMs) are widely marketed for their ability to solve complex tasks that supposedly require \textit{reasoning}. However, it is still debated whether LMs engage in structured reasoning or whether they are merely replicating statistical relations from the data on which they are trained \citep{benderDangersStochasticParrotsCanLanguage2021, huangReasoningLarge2023, mirzadehGSMSymbolicUnderstandingLimitationsMathematicalReasoning2024, kambhampati_stop_2025}. Flagship models developed by companies such as OpenAI, Anthropic, and Alibaba are labeled as \textit{reasoning} models that generate long ``chains of thought'' before generating the final output. Models from the Qwen-series output these traces in designated $<think>$ tags \citep{yang_qwen3_2025}.

In this paper we assess standard definitions of reasoning and the current framing of reasoning in the NLP literature, and argue that the definitions of reasoning are incommensurable with transformer computations. The use of ill-fitting terminology is problematic due to connotations attached to a notion such as \textit{reasoning}, which has a rich tradition in fields such as philosophy, AI, and psychology \citep[see e.g.][]{tversky_judgment_1974, tversky_framing_1981, tversky_rational_1986, albusOutlineTheoryIntelligence1991, wallacePracticalReason2024, jiang_peek_2024}. \textit{Reasoning} is a key criteria for a system to be considered intelligent, and it is therefore a crucial aspect of the scientific aims pursued in the field of \textit{AI} \citep{hojer_research_2025}. 

We describe LMs as \textit{implicit Markov kernels} similar to \citet{zekri_large_2025}, but use this formalization to frame a discussion of the notion of \textit{reasoning} and \textit{inference}. This lens challenges the description of transformer computations as commensurable with the process of ``thinking in a logical and systematic manner''. Multiple papers have experimentally illustrated the logical shortcomings of LMs by showing the failure-modes when it comes to simple ``\textit{reasoning}'' tasks \citep{nezhurinaAliceWonderlandSimpleTasksShowing2024, mirzadehGSMSymbolicUnderstandingLimitationsMathematicalReasoning2024, jiang_peek_2024}.

%% file: tex/02-reasoning.tex
\section{Reasoning and Natural Language Processing}

\textit{Reasoning} is increasingly being used to describe a specific type of output generated by LMs, namely the ``thought traces'' as they have been labeled after \citet{wei2023} and earlier work by \citet{nye_show_2021}. Concurrently, it is being debated whether LMs can be said to be doing anything akin to \textit{reasoning}. A key issue seems to be that the grounds for disagreement are not necessarily clear. We argue that addressing the well-formedness of the question of \textit{reasoning} in LMs is of importance to \textit{AI} research, while others may claim that it is enough that a system seems to be \textit{reasoning} if it results in more correct outputs. We return to this in \autoref{sec:discussion}.

\subsection{Definitions of Reasoning}

It is not that there are no definitions of \textit{reasoning}. To reason about the natural world has a history which is documented as far back as to the ancient Greeks, and while the history of \textit{reasoning} is fundamental to science, it is beyond the scope of this paper to lay it out in detail. Surveying definitions can be summarized to \textit{reasoning} being something along the lines of ``\textit{thinking about something in a logical manner}'' or ``\textit{the process of thinking about something logically and systematically to make an inference}''.
In psychology and philosophy one finds definitions that are not too far off from these statements (see e.g. \citet{johnson-laird_how_2008, johnson-lairdMentalModelsHumanReasoning2010, over_human_2024, albusOutlineTheoryIntelligence1991, portoraroAutomatedReasoning2025}). Of special interest to the ongoing discussion in NLP and \textit{AI}, \citet{tversky_framing_1981, tversky_rational_1986} discussed \textit{the principle of invariance} in the theory of rational decision-making to illustrate that a rational decision-maker should be invariant to various biases and logical fallacies.

If we look at its use in the NLP literature, the term has been increasing rather rapidly over the past five years. We trace this narrative in key papers from the NLP literature that have been influential for the literature on \textit{reasoning} (\autoref{sec:timeline}).

\citet{huangReasoningLarge2023} surveyed recent papers on LM reasoning. They define \textit{reasoning} both as a ``cognitive process that involves using evidence, arguments, and logic to arrive at conclusions or make judgment'' and as ``the process of thinking about something in a logical and systematic way, using evidence and past experiences to reach a conclusion or make a decision''. Later they demarcate between \textit{formal reasoning}, which is a ``systematic and logical process that follows a set of rules and principles (...)'', and \textit{informal reasoning}, which is ``a less structured approach that relies on intuition, experience, and common sense to draw conclusions and solve problems (...)''. It should be clear to the reader that \textbf{what is meant by \textit{reasoning} is not clear at all}.

\subsection{A Timeline of LM Reasoning}\label{sec:timeline}

The labeling of autoregressive LM generation as \textit{reasoning} has arisen in the last decade. \citet{vaswaniAttentionAllYouNeed2023} introduced the transformer with self-attention, enabling efficient scaling of language models in terms of both model parameters and dataset size. \citet{radford_language_2019} then introduced GPT-2 calling it a multi-task learner, referring to the finding that the model was capable of solving a vast variety of traditional NLP tasks that one would usually have to fine-tune a specific model for. \citet{brownLanguageModelsAreFewShotLearners2020} introduced the follow-up model, GPT-3, as a few-shot learner - namely one that is capable of learning by example. Research on the capabilities of these models followed soon after with \citet{wei2023} arguing that Chain-of-Thought (CoT) prompting elicits \textit{reasoning} in LMs. \citet{kojima_large_2023} argue that LMs are not only few-shot reasoners, but that they are zero-shot reasoners and \citet{wang_self-consistency_2023} show that self-consistency improves CoT \textit{reasoning} in LMs. This timeline indicates how ``reasoning'' has become a function of the format of the model output as opposed to a logical process. It has become evident how important the \textit{right} data is for enabling ``reasoning''. Early examples hereof are the \textit{TinyStories} models which were trained with fully synthetic data \citep{eldan_tinystories_2023}, the Phi-family of models that are trained on especially curated datasets to excel at python coding tasks and other forms of ``reasoning'' tasks \citep{gunasekar_textbooks_2023, li_textbooks_2023}, and the Orca models which were trained specifically on data containing ``reasoning'' strategies generated by GPT-4 \citep{mukherjee_orca_2023, mitra_orca_2023}. More recently we have seen models that are trained initially on extensive examples of ``reasoning'' and then optimized using various forms of reinforcement learning to incentivize the generation of long ``reasoning'' traces. \cite{guoDeepSeekR1IncentivizesReasoningLLMsReinforcement2025} provide the technical details of this process. They furthermore claim that reasoning \textit{emerges} spontaneously from the training procedures and \citet{weiEmergentAbilitiesLargeLanguageModels2022a} and \citet{bowman_eight_2023} claim that certain \textit{behaviors} \textit{emerge} unpredictably.


\subsection{Logic and Large language Models}

The autoregressive aspect and relatedly the finding that CoT improves the performance of an LM is used to argue that LMs \textit{reason} over data \citep{brownLanguageModelsAreFewShotLearners2020, wei2023, kojima_large_2023}. But it is important to consider the fact that autoregressive generation applies the same model $\kappa_{\theta}$ iteratively in the generation of new tokens as illustrated below. Assuming no sampling during generation one would get the exact same output sequence when prompting an LM with $x_1$ (``The cat'') and $x_2$ (``The cat sat'') if $\kappa_{\theta}(\text{``sat''} | \text{``the cat''}) \geq \kappa_{\theta}(x | \text{``the cat ''})$ for all $x \in \Sigma$. When a different sampling technique is applied the diagram would appear to be branching.

\begin{center}
\begin{tikzpicture}[node distance=2.4cm,>=latex]
    \node[state] (x0) {$x_0$};
    \node[state,right of=x0] (x1) {$x_1$};
    \node[state,right of=x1] (x2) {$x_2$};
    \node[state,right of=x2] (x3) {$...$};
    \node[state,right of=x3] (xn) {$x_n$};
    
    \draw[->] (x0) -- (x1) node[midway,above] {$\kappa_\theta$};
    \draw[->] (x1) -- (x2) node[midway,above] {$\kappa_\theta$};
    \draw[->] (x2) -- (x3) node[midway,above] {$\kappa_\theta$};
    \draw[->] (x3) -- (xn) node[midway,above] {$\kappa_\theta$};
    
    \node[below=0.3cm of x0,align=center,font=\small]
    {``The''};
    \node[below=0.3cm of x1,align=center,font=\small]
    {``The cat''};
    \node[below=0.3cm of x2,align=center,font=\small]
    {``The cat sat''};
    \node[below=0.3cm of x3,align=center,font=\small]
    {``The cat sat on''};
    \node[below=0.3cm of xn,align=center,font=\small]
    {``The ... '' [EOS]};
\end{tikzpicture}
\end{center}

We observe a narrative in the discourse on NLP and AI where LMs are framed as general-purpose technologies capable of \textit{reasoning} and generalizing to almost any type of text-based problem (see \autoref{sec:timeline}). However, it is also clear from this literature that the ``reasoning'' is strictly a feature of the format of the model outputs and not related to the logic of the system generating the outputs - although definitions imply otherwise.

It is also clear from the definitions that logic and systematicity are key factors of \textit{reasoning}. We thus ask: are LMs logical and systematic? If one wishes to state that an LM \textit{reasons} and generates logical statements, the logic that is enforced would \textbf{only} be the logic of the data, so to speak. However, it is not known whether LMs can --- in principle --- enforce any logical structures, and recent research displays evidence to the contrary \citep{peng_limitations_2024}. As an example and to illustrate our point we describe an LM formally as a \textbf{Markov process}.

%% file: tex/03-markov.tex
\section{Language Models and Markov Processes}

To illustrate why LMs do not \textit{reason} in the sense of the definitions emphasized in this paper, it can be instructive to view LMs through a different lens and revisit the fundamentals. Shannon explains in his 1948 paper:
\begin{quote}
    [A discrete source] will choose successive symbols according to certain probabilities depending, in general, on preceding choices as well as the particular symbols in question. (...) a mathematical model of a system which produces such a sequence of symbols governed by a set of probabilities, is known as a stochastic process. \citep[][p. 5]{shannonMathematicalTheory1948}.
\end{quote}

\subsection{Markov processes}

The Markov property requires that a stochastic process is \textit{memoryless}, meaning that its future states are entirely dependent on the current state and not on its history \citep{murphy_probabilistic_2022}. For a fixed maximum context length $L$ and finite vocabulary $V$, an autoregressive LM with the state $s_t = x_{t-L:t-1}$ induces an \textit{order}-$L$ Markov chain over $V^L$.

A \textit{Markov kernel} is a measurable mapping $\kappa : X \to \mathcal{P}(Y)$, that assigns to each input $x \in X$ a probability distribution over possible outputs in $Y$. Thus, for any context $x$, the kernel $\kappa(\cdot \mid x)$ defines the conditional probability distribution of outcomes. An LM is a parameterized kernel $\kappa_{\theta}$. An invariance in a kernel refers to transformations of the input space that leave the output distribution unchanged. For example, if two contexts are approximately logically equivalent, one might expect the kernel to assign (almost) the same continuation distribution:
$$
x_i \approx x_j \;\Rightarrow\; \kappa_{\theta}(\cdot \mid x_i) \approx \kappa_{\theta}(\cdot \mid x_j).
$$

In an LM the kernel is implicit. It is not defined explicitly as a transition matrix, but is instead instantiated by the parameters of the model $\kappa_{\theta}$. The training objective maximizes the likelihood of observed sequences by minimizing the cross-entropy between the empirical data distribution on which the model is trained and the model distribution. This learning objective thus approximates the empirical conditional distribution of tokens. Specifically, an LM is a Markov model on a finite state space as both the vocabulary and context of a model are finite \citep{zekri_large_2025}. 

But this objective does not enforce global invariances or logical implications, although the window of what is local is widening as models are trained with larger contexts. At most, it \textit{loosely enforces strong regularities in the data} as invariance. When an LM ``reasons'' and applies a certain logic it corresponds to regularities in the kernel.

Crucially, even approximate invariance is not guaranteed and LMs often violate logical consistency \citep{nezhurinaAliceWonderlandSimpleTasksShowing2024, mirzadehGSMSymbolicUnderstandingLimitationsMathematicalReasoning2024, shojaee_illusion_2025}. This illustrates the fact that invariances in transformer kernels are statistical artifacts of the training data distribution and not structural properties of the architecture or learning objectives that enforce a logic on model outputs. This is evidenced by how important data has become for training models that are better at ``reasoning'' as discussed in \autoref{sec:timeline}.

\subsection{``Logic'' as Approximate Invariances in the Kernel}

Rather than examining \textit{reasoning} as a logical process, researchers usually measure how well a model adheres to logical and deductive implications \citep{nezhurinaAliceWonderlandSimpleTasksShowing2024, mirzadehGSMSymbolicUnderstandingLimitationsMathematicalReasoning2024, mondorfAccuracyEvaluating2024}.

In this context, an invariance in a parameterized Markov kernel $\kappa_{\theta}$ refers to the property that certain transformations of input contexts $x$ lead to approximately the same conditional output distribution. Approximate invariances of the kernel can be viewed as the model’s implicit expression of epistemic stability, and we can interpret \textit{reasoning}-like consistency as such. Similar, but different, contexts that have similar ``structure'' should map to similar continuation distributions.

Let $V$ be a finite vocabulary and $X = V^{\leq L}$ be the space of token sequences up to a context window of size $L$. An LM parameterized by $\theta$ then defines a Markov kernel
$$
\kappa_{\theta} : X \to \Delta(V), \quad \kappa_{\theta}(\cdot \mid x) = p_{\theta}(x_t \mid x_{t-L:t-1}),
$$
which maps a bounded context $x$ to the probability simplex over the vocabulary $\Delta(V)$. As $L$ is finite the model is an \textit{order}-$L$ Markov chain.

In this setting, ``reasoning'' is the ability to preserve inferential relations between symbolic structures across context, which is viewed as a property of \textit{invariance} of the kernel given a transformation of the input. One might then distinguish between two related types of invariances: \textit{transformation invariances} and \textit{inferential invariances}.

Let $T$ be a group of logic-preserving transformations. A model is \textit{approximately transformation-invariant} under $T$ if
$$
\forall\, t \in T,\quad \mathbf{V_T}\!\big(\kappa_{\theta}(\cdot \mid x),\, \kappa_{\theta}(\cdot \mid t(x))\big) \le \epsilon_T,
$$
where $\mathbf{V_T}$ denotes total variation distance between distributions on $V$ for a group of transformations $T$. In words, an LM's predictions should not change substantially under transformations that preserve logic.

Suppose now we have a logical reasoning rule $r$ (such as \textit{modus ponens}); define a relation $\mathcal{R}_r = \{(x, y): x \xRightarrow{r} y\}$. A model exhibits \textit{inferential invariance} with respect to a rule $r$ if
$$
\forall\, (x, y) \in \mathcal{R}_r,\quad \kappa_{\theta}(y \mid x) \ge 1 - \delta_r,
$$
and the relation holds under the aforementioned transformations. What do these measures mean conceptually? $\epsilon_T$ and $\delta_r$ are an attempt to capture the \textit{approximate invariances} of a model. An $\epsilon_T > 0$ indicates that a model is somewhat sensitive to irrelevant transformations and $\delta_r > 0$ indicates the degree to which ``reasoning'' is imperfect. LMs are usually said not to \textit{reason} because they are not invariant to irrelevant transformations on $x$ \citep[e.g.][]{mirzadehGSMSymbolicUnderstandingLimitationsMathematicalReasoning2024}.

LMs are trained to match an empirical probability distribution of a dataset $\mathcal{D} \subset X$ and do not directly optimize for, or enforce, these invariances. They instead reproduce regularities in $\mathcal{D}$ that approximate inferential structure. If a change in model output results in a different semantic interpretation, the apparent \textit{reasoning} surely reflects data-induced regularities rather than logical inference.

To make research into transformation invariance directly comparable between papers, notions such as these could be operationalised and applied to measure epistemic fidelity in papers such as \citet{nezhurinaAliceWonderlandSimpleTasksShowing2024, mirzadehGSMSymbolicUnderstandingLimitationsMathematicalReasoning2024, jiang_peek_2024}. However, we emphasize that this would do no more than illustrate the stability of $\kappa_{\theta}$.

\subsection{Inference}

Now, we postulate that we could change almost every mention of ``reasoning'' in this paper with \textbf{inference}. Inference does not carry the psychological connotations of a term such as \textit{reasoning} and is well-defined both within statistics and machine learning. In statistics \textbf{inference} is passing from sample data to a generalization, and it is usually equivalent to the notion of \textit{prediction} in machine learning \citep{murphy_probabilistic_2022}.

In no sense is what we have discussed as ``reasoning'' different from the notion of \textit{inference}, begging the question of why researchers speak of \textit{reasoning} in a formal science. From a purely structuralist perspective, the view that words get their meaning from the contexts in which they are used, this would indicate that \textit{reasoning} $\approx$ \textit{inference}. But this would be false by the definitions we have presented in this paper. Analyzing LM operations should therefore be kept a science of systematic natural language \textbf{inference} and not one of \textit{reasoning}. Furthermore, this framing makes the object of analysis clearer: it is not about imbuing \textit{reasoning} but about ensuring logical and systematic \textit{inference}.

\subsection{Research Program}

Empirical research can help elucidate how epistemic uncertainty manifests as invariance in model predictions when logical structures in the data are controlled. We aim to show that ``reasoning'' in LMs should rather be framed as \textbf{inference}, and to illustrate how the proposed metrics vary when certain logical structures (regularities in $\mathcal{D}$) are enforced in the datasets used to train LMs. We aim to build simple synthetic datasets wherein we control the logical structure of tokens; with these datasets we then train small toy transformers to establish how the theory generalizes to the transformer architecture, which computes $\Delta(V)$ quite differently from a discrete Markov model.

It is clear that generating longer CoT traces results in better benchmark performance. But this is a result of more expressive models due to a higher parameter count and CoT (as shown by \citet{merrill_expressive_2024}), and thus more accurate \textbf{inference} over the distribution of data on which an LM is trained. \citet{stechly_beyond_2025} showed that even semantically invalid CoT traces lead to better performance on certain tasks. Based on this we additionally aim to analyze the operations of autoregressive LM \textbf{inference} empirically to elucidate the logic of such a result.

%% file: tex/04-discussion.tex
\section{Discussion}\label{sec:discussion}

The discrete Markov models discussed in this paper are simple in terms of their expressive power when compared to modern LMs. However, most LMs can theoretically be framed as Markov chains \citep{zekri_large_2025}. This provides fertile ground for theorizing about the nature of such models; in our case, specifically whether they can be said to \textit{reason} given current definitions of \textit{reasoning}. And furthermore, whether that is the right question to ask; formulating the right question is not separate from the science, it is part of it. Conflating \textit{reasoning} and \textbf{inference} can obscure how epistemic uncertainty is represented, and thus sticking with clear definitions and terminology enables researchers to be rigid.

The implicit nature of the kernel of an LM is not insignificant as it obscures the structure of the kernel. In an LM the input is embedded via a learned embedding function $e^{in}: X \to \mathbb{R}^d$, modified by a positional embedding, and then transformed iteratively by a combination of an attention-mechanism and a feed-forward block before the application of a so-called ``de-embedding'' function $e^{out}: \mathbb{R}^d \to X$ \citep{vaswaniAttentionAllYouNeed2023}. These operations are done in a real-valued continuous space with induced non-linearities imbuing a certain structure on the space of the operations.

We ask: do (and should) these added complexities fundamentally change the conceptualization of the operations computed by a model? Neural networks trained for extracting word embeddings have e.g. been proven to approximate a matrix factorization of the PMI between terms in a corpus; a factorization problem to which there is a unique solution, namely the singular value decomposition \citep{levy_neural_2014, goldberg_word2vec_2014}. While it has been shown that the attention mechanism increases expressivity and that CoT generation makes a model strictly more expressive \citep{merrill_expressive_2024}, it is nonetheless a model optimizing a loss-function that yields the most likely next token given a corpus.

With this paper we aim to ground the debate of logical \textit{reasoning} in LMs in terms of invariances in the learned kernel as opposed to logical and systematic \textit{thinking}. If we can reduce \textit{reasoning} to \textbf{inference} and something like a description of invariances in a kernel, we have a solid foundation for demystifying and discussing the capabilities and limitations of an LM.

%% file: tex/05-conclusion.tex
\section{Conclusion}

In conclusion, we reiterate and emphasize the result that an LM can be seen as implementing an \textbf{implicit Markov kernel}. Based on this result and an investigation of current use of the notion of \textit{reasoning}, we argue that \textit{reasoning} in LMs is in no meaningful way separate from the notion of \textbf{inference}. Framing reasoning as inference under epistemic uncertainty refocuses the debate from vague notions of \textit{reasoning} to measurable epistemic properties of model-based inference.
We formalize simple metrics of \textit{invariance}, discuss how they relate to the notion of \textit{reasoning} in current research and suggest a research program to understand the operations of LMs that could enable \textit{logical inference} while also elucidating the limitations of LMs.

%% file: references.bib
@online{nezhurinaAliceWonderlandSimpleTasksShowing2024,
  title = {Alice in {{Wonderland}}: {{Simple Tasks Showing Complete Reasoning Breakdown}} in {{State-Of-the-Art Large Language Models}}},
  shorttitle = {Alice in {{Wonderland}}},
  author = {Nezhurina, Marianna and Cipolina-Kun, Lucia and Cherti, Mehdi and Jitsev, Jenia},
  date = {2024-07-13},
  year = {2024},
  eprint = {2406.02061},
  eprinttype = {arXiv},
  eprintclass = {cs},
  
  url = {http://arxiv.org/abs/2406.02061},
  urldate = {2025-02-04},
  pubstate = {prepublished}
}

@misc{kojima_large_2023,
    title = {Large {Language} {Models} are {Zero}-{Shot} {Reasoners}},
    url = {http://arxiv.org/abs/2205.11916},
    urldate = {2025-03-27},
    publisher = {arXiv},
    author = {Kojima, Takeshi and Gu, Shixiang Shane and Reid, Machel and Matsuo, Yutaka and Iwasawa, Yusuke},
    month = jan,
    year = {2023},
}

@article{radford_language_2019,
    title = {Language {Models} are {Unsupervised} {Multitask} {Learners}},
    language = {en},
    author = {Radford, Alec and Wu, Jeffrey and Child, Rewon and Luan, David and Amodei, Dario and Sutskever, Ilya},
    year = {2019},}

@misc{brownLanguageModelsAreFewShotLearners2020,
    title = {Language {Models} are {Few}-{Shot} {Learners}},
    url = {http://arxiv.org/abs/2005.14165},
    urldate = {2024-04-25},
    publisher = {arXiv},
    author = {Brown, Tom B. and Mann, Benjamin and Ryder, Nick and Subbiah, Melanie and Kaplan, Jared and Dhariwal, Prafulla and Neelakantan, Arvind and Shyam, Pranav and Sastry, Girish and Askell, Amanda and Agarwal, Sandhini and Herbert-Voss, Ariel and Krueger, Gretchen and Henighan, Tom and Child, Rewon and Ramesh, Aditya and Ziegler, Daniel M. and Wu, Jeffrey and Winter, Clemens and Hesse, Christopher and Chen, Mark and Sigler, Eric and Litwin, Mateusz and Gray, Scott and Chess, Benjamin and Clark, Jack and Berner, Christopher and McCandlish, Sam and Radford, Alec and Sutskever, Ilya and Amodei, Dario},
    month = jul,
    year = {2020},
}

@misc{wei2023,
    title = {Chain-of-{Thought} {Prompting} {Elicits} {Reasoning} in {Large} {Language} {Models}},
    url = {http://arxiv.org/abs/2201.11903},
    urldate = {2023-10-24},
    publisher = {arXiv},
    author = {Wei, Jason and Wang, Xuezhi and Schuurmans, Dale and Bosma, Maarten and Ichter, Brian and Xia, Fei and Chi, Ed and Le, Quoc and Zhou, Denny},
    month = jan,
    year = {2023},
}

@misc{wang_self-consistency_2023,
    title = {Self-{Consistency} {Improves} {Chain} of {Thought} {Reasoning} in {Language} {Models}},
    url = {http://arxiv.org/abs/2203.11171},
    urldate = {2025-04-06},
    publisher = {arXiv},
    author = {Wang, Xuezhi and Wei, Jason and Schuurmans, Dale and Le, Quoc and Chi, Ed and Narang, Sharan and Chowdhery, Aakanksha and Zhou, Denny},
    month = mar,
    year = {2023},
}

@article{guoDeepSeekR1IncentivizesReasoningLLMsReinforcement2025,
  title = {{{DeepSeek-R1}} Incentivizes Reasoning in {{LLMs}} through Reinforcement Learning},
  author = {Guo, Daya and Yang, Dejian and Zhang, Haowei and Song, Junxiao et. al.},
  date = {2025-09},
  year = {2025},
  journaltitle = {Nature},
  volume = {645},
  number = {8081},
  pages = {633--638},
  publisher = {Nature Publishing Group},
  issn = {1476-4687},
  
  url = {https://www.nature.com/articles/s41586-025-09422-z},
  urldate = {2025-10-10},
  langid = {english}
}

@inproceedings{huangReasoningLarge2023,
    address = {Toronto, Canada},
    title = {Towards {Reasoning} in {Large} {Language} {Models}: {A} {Survey}},
    shorttitle = {Towards {Reasoning} in {Large} {Language} {Models}},
    url = {https://aclanthology.org/2023.findings-acl.67},
    urldate = {2023-11-13},
    booktitle = {Findings of the {Association} for {Computational} {Linguistics}: {ACL} 2023},
    publisher = {Association for Computational Linguistics},
    author = {Huang, Jie and Chang, Kevin Chen-Chuan},
    editor = {Rogers, Anna and Boyd-Graber, Jordan and Okazaki, Naoaki},
    month = jul,
    year = {2023},
    
}

@misc{weiEmergentAbilitiesLargeLanguageModels2022a,
    title = {Emergent {Abilities} of {Large} {Language} {Models}},
    url = {https://arxiv.org/abs/2206.07682v2},
    language = {en},
    urldate = {2024-10-08},
    journal = {arXiv.org},
    author = {Wei, Jason and Tay, Yi and Bommasani, Rishi and Raffel, Colin and Zoph, Barret and Borgeaud, Sebastian and Yogatama, Dani and Bosma, Maarten and Zhou, Denny and Metzler, Donald and Chi, Ed H. and Hashimoto, Tatsunori and Vinyals, Oriol and Liang, Percy and Dean, Jeff and Fedus, William},
    month = jun,
    year = {2022}}

@misc{bowman_eight_2023,
    title = {Eight {Things} to {Know} about {Large} {Language} {Models}},
    url = {http://arxiv.org/abs/2304.00612},
    urldate = {2025-10-04},
    publisher = {arXiv},
    author = {Bowman, Samuel R.},
    month = apr,
    year = {2023},
}

@misc{mirzadehGSMSymbolicUnderstandingLimitationsMathematicalReasoning2024,
    title = {{GSM}-{Symbolic}: {Understanding} the {Limitations} of {Mathematical} {Reasoning} in {Large} {Language} {Models}},
    shorttitle = {{GSM}-{Symbolic}},
    url = {http://arxiv.org/abs/2410.05229},
    urldate = {2024-10-14},
    publisher = {arXiv},
    author = {Mirzadeh, Iman and Alizadeh, Keivan and Shahrokhi, Hooman and Tuzel, Oncel and Bengio, Samy and Farajtabar, Mehrdad},
    month = oct,
    year = {2024},
}

@article{shojaee_illusion_2025,
    title = {The {Illusion} of {Thinking}: {Understanding} the {Strengths} and {Limitations} of {Reasoning} {Models} via the {Lens} of {Problem} {Complexity}},
    language = {en},
    year = {2025},
    author = {Shojaee, Parshin and Mirzadeh, Iman and Alizadeh, Keivan and Horton, Maxwell and Bengio, Samy and Farajtabar, Mehrdad}}

@misc{yang_qwen3_2025,
    title = {Qwen3 {Technical} {Report}},
    url = {http://arxiv.org/abs/2505.09388},
    urldate = {2025-06-03},
    publisher = {arXiv},
    author = {Yang, An and Li, Anfeng and Yang, Baosong and Zhang, Beichen and Hui, Binyuan and Zheng, Bo et. al.},
    month = may,
    year = {2025},
}

@inproceedings{hojer_research_2025,
    address = {Vienna},
    title = {Research {Community} {Perspectives} on ``{Intelligence}'' and {Large} {Language} {Models}},
    language = {English},
    booktitle = {The {Findings} of the {Association} of {Computations} {Linguistics}},
    author = {Højer, Bertram and Jakobsen, Terne Thorn and Rogers, Anna and Heinrich, Stefan},
    year = {2025}}

@misc{kambhampati_stop_2025,
    title = {Stop {Anthropomorphizing} {Intermediate} {Tokens} as {Reasoning}/{Thinking} {Traces}!},
    url = {http://arxiv.org/abs/2504.09762},
    urldate = {2025-06-06},
    publisher = {arXiv},
    author = {Kambhampati, Subbarao and Stechly, Kaya and Valmeekam, Karthik and Saldyt, Lucas and Bhambri, Siddhant and Palod, Vardhan and Gundawar, Atharva and Samineni, Soumya Rani and Kalwar, Durgesh and Biswas, Upasana},
    month = may,
    year = {2025},
}

@article{albusOutlineTheoryIntelligence1991,
    title = {Outline for a theory of intelligence},
    volume = {21},
    copyright = {https://ieeexplore.ieee.org/Xplorehelp/downloads/license-information/IEEE.html},
    issn = {00189472},
    url = {http://ieeexplore.ieee.org/document/97471/},
    language = {en},
    number = {3},
    urldate = {2025-02-13},
    journal = {IEEE Transactions on Systems, Man, and Cybernetics},
    author = {Albus, J.S.},
    month = jun,
    year = {1991},
    
}

@incollection{wallacePracticalReason2024,
  title = {Practical {{Reason}}},
  booktitle = {The {{Stanford Encyclopedia}} of {{Philosophy}}},
  author = {Wallace, R. Jay and Kiesewetter, Benjamin},
  editor = {Zalta, Edward N. and Nodelman, Uri},
  year = {2024},
  edition = {Fall 2024},
  publisher = {Metaphysics Research Lab, Stanford University},
  url = {https://plato.stanford.edu/archives/fall2024/entries/practical-reason/},
  urldate = {2025-04-04}
}

@article{shannonMathematicalTheory1948,
    title = {A {Mathematical} {Theory} of {Communication}},
    volume = {Vol. 27},
    language = {en},
    journal = {The Bell System Technical Journal},
    author = {Shannon, C E},
    year = {1948},
    
}

@inproceedings{levy_neural_2014,
    address = {Cambridge, MA, USA},
    series = {{NIPS}'14},
    title = {Neural word embedding as implicit matrix factorization},
    volume = {2},
    urldate = {2025-09-23},
    booktitle = {Proceedings of the 28th {International} {Conference} on {Neural} {Information} {Processing} {Systems} - {Volume} 2},
    publisher = {MIT Press},
    author = {Levy, Omer and Goldberg, Yoav},
    month = dec,
    year = {2014},
    
}

@misc{goldberg_word2vec_2014,
    title = {word2vec {Explained}: deriving {Mikolov} et al.'s negative-sampling word-embedding method},
    shorttitle = {word2vec {Explained}},
    url = {http://arxiv.org/abs/1402.3722},
    urldate = {2025-08-07},
    publisher = {arXiv},
    author = {Goldberg, Yoav and Levy, Omer},
    month = feb,
    year = {2014},
}

@inproceedings{benderDangersStochasticParrotsCanLanguage2021,
    address = {Virtual Event Canada},
    title = {On the {Dangers} of {Stochastic} {Parrots}: {Can} {Language} {Models} {Be} {Too} {Big}?},
    isbn = {978-1-4503-8309-7},
    shorttitle = {On the {Dangers} of {Stochastic} {Parrots}},
    url = {https://dl.acm.org/doi/10.1145/3442188.3445922},
    
    language = {en},
    urldate = {2024-09-18},
    booktitle = {Proceedings of the 2021 {ACM} {Conference} on {Fairness}, {Accountability}, and {Transparency}},
    publisher = {ACM},
    author = {Bender, Emily M. and Gebru, Timnit and McMillan-Major, Angelina and Shmitchell, Shmargaret},
    month = mar,
    year = {2021},
    
}

@misc{eldan_tinystories_2023,
    title = {{TinyStories}: {How} {Small} {Can} {Language} {Models} {Be} and {Still} {Speak} {Coherent} {English}?},
    shorttitle = {{TinyStories}},
    url = {http://arxiv.org/abs/2305.07759},
    urldate = {2025-10-09},
    publisher = {arXiv},
    author = {Eldan, Ronen and Li, Yuanzhi},
    month = may,
    year = {2023},
}

@misc{gunasekar_textbooks_2023,
    title = {Textbooks {Are} {All} {You} {Need}},
    url = {http://arxiv.org/abs/2306.11644},
    urldate = {2025-10-10},
    publisher = {arXiv},
    author = {Gunasekar, Suriya and Zhang, Yi and Aneja, Jyoti and Mendes, Caio César Teodoro and Giorno, Allie Del and Gopi, Sivakanth and Javaheripi, Mojan and Kauffmann, Piero and Rosa, Gustavo de and Saarikivi, Olli and Salim, Adil and Shah, Shital and Behl, Harkirat Singh and Wang, Xin and Bubeck, Sébastien and Eldan, Ronen and Kalai, Adam Tauman and Lee, Yin Tat and Li, Yuanzhi},
    month = oct,
    year = {2023},
}

@misc{li_textbooks_2023,
    title = {Textbooks {Are} {All} {You} {Need} {II}: phi-1.5 technical report},
    shorttitle = {Textbooks {Are} {All} {You} {Need} {II}},
    url = {http://arxiv.org/abs/2309.05463},
    urldate = {2025-10-10},
    publisher = {arXiv},
    author = {Li, Yuanzhi and Bubeck, Sébastien and Eldan, Ronen and Giorno, Allie Del and Gunasekar, Suriya and Lee, Yin Tat},
    month = sep,
    year = {2023},
}

@misc{mukherjee_orca_2023,
    title = {Orca: {Progressive} {Learning} from {Complex} {Explanation} {Traces} of {GPT}-4},
    shorttitle = {Orca},
    url = {http://arxiv.org/abs/2306.02707},
    urldate = {2025-10-10},
    publisher = {arXiv},
    author = {Mukherjee, Subhabrata and Mitra, Arindam and Jawahar, Ganesh and Agarwal, Sahaj and Palangi, Hamid and Awadallah, Ahmed},
    month = jun,
    year = {2023},
}

@misc{mitra_orca_2023,
    title = {Orca 2: {Teaching} {Small} {Language} {Models} {How} to {Reason}},
    shorttitle = {Orca 2},
    url = {http://arxiv.org/abs/2311.11045},
    urldate = {2025-10-10},
    publisher = {arXiv},
    author = {Mitra, Arindam and Corro, Luciano Del and Mahajan, Shweti and Codas, Andres and Simoes, Clarisse and Agarwal, Sahaj and Chen, Xuxi and Razdaibiedina, Anastasia and Jones, Erik and Aggarwal, Kriti and Palangi, Hamid and Zheng, Guoqing and Rosset, Corby and Khanpour, Hamed and Awadallah, Ahmed},
    month = nov,
    year = {2023},
}

@misc{zekri_large_2025,
    title = {Large {Language} {Models} as {Markov} {Chains}},
    url = {http://arxiv.org/abs/2410.02724},
    urldate = {2025-10-10},
    publisher = {arXiv},
    author = {Zekri, Oussama and Odonnat, Ambroise and Benechehab, Abdelhakim and Bleistein, Linus and Boullé, Nicolas and Redko, Ievgen},
    month = feb,
    year = {2025},
}

@book{johnson-laird_how_2008,
    title = {How {We} {Reason}},
    isbn = {978-0-19-955133-0},
    url = {https://academic.oup.com/book/11984},
    urldate = {2024-09-12},
    publisher = {Oxford University Press},
    author = {Johnson-Laird, Philip},
    month = oct,
    year = {2008},
    
    
}

@article{johnson-lairdMentalModelsHumanReasoning2010,
    title = {Mental models and human reasoning},
    url = {https://www.pnas.org/doi/10.1073/pnas.1012933107},
    
    language = {en},
    urldate = {2025-03-27},
    author = {Johnson-Laird,, Philip},
    year = {2010},
    
}

@book{over_human_2024,
    address = {Cambridge New York (N.Y.)},
    series = {Elements in philosophy of mind},
    title = {Human reasoning},
    isbn = {978-1-009-49531-8},
    language = {eng},
    publisher = {Cambridge University press},
    author = {Over, David E. and Evans, Jonathan St B. T.},
    year = {2024},
    
}

@misc{vaswaniAttentionAllYouNeed2023,
    title = {Attention {Is} {All} {You} {Need}},
    url = {http://arxiv.org/abs/1706.03762},
    urldate = {2024-10-03},
    publisher = {arXiv},
    author = {Vaswani, Ashish and Shazeer, Noam and Parmar, Niki and Uszkoreit, Jakob and Jones, Llion and Gomez, Aidan N. and Kaiser, Lukasz and Polosukhin, Illia},
    month = aug,
    year = {2017},
}

@inproceedings{jiang_peek_2024,
    address = {Miami, Florida, USA},
    title = {A {Peek} into {Token} {Bias}: {Large} {Language} {Models} {Are} {Not} {Yet} {Genuine} {Reasoners}},
    shorttitle = {A {Peek} into {Token} {Bias}},
    url = {https://aclanthology.org/2024.emnlp-main.272/},
    urldate = {2025-10-12},
    booktitle = {Proceedings of the 2024 {Conference} on {Empirical} {Methods} in {Natural} {Language} {Processing}},
    publisher = {Association for Computational Linguistics},
    author = {Jiang, Bowen and Xie, Yangxinyu and Hao, Zhuoqun and Wang, Xiaomeng and Mallick, Tanwi and Su, Weijie J and Taylor, Camillo Jose and Roth, Dan},
    editor = {Al-Onaizan, Yaser and Bansal, Mohit and Chen, Yun-Nung},
    month = nov,
    year = {2024},
    
    pages = {4722--4756},
}

@article{tversky_framing_1981,
    title = {The {Framing} of {Decisions} and the {Psychology} of {Choice}},
    volume = {211},
    language = {en},
    author = {Tversky, Amos and Kahneman, Daniel},
    year = {1981},
    
}

@article{tversky_rational_1986,
    title = {Rational {Choice} and the {Framing} of {Decisions}},
    volume = {59},
    issn = {0021-9398},
    url = {https://www.jstor.org/stable/2352759},
    number = {4},
    urldate = {2025-10-13},
    journal = {The Journal of Business},
    author = {Tversky, Amos and Kahneman, Daniel},
    year = {1986},
    
    pages = {S251--S278},
}

@misc{nye_show_2021,
    title = {Show {Your} {Work}: {Scratchpads} for {Intermediate} {Computation} with {Language} {Models}},
    shorttitle = {Show {Your} {Work}},
    url = {http://arxiv.org/abs/2112.00114},
    urldate = {2025-10-14},
    publisher = {arXiv},
    author = {Nye, Maxwell and Andreassen, Anders Johan and Gur-Ari, Guy and Michalewski, Henryk and Austin, Jacob and Bieber, David and Dohan, David and Lewkowycz, Aitor and Bosma, Maarten and Luan, David and Sutton, Charles and Odena, Augustus},
    month = nov,
    year = {2021},
}

@misc{merrill_expressive_2024,
    title = {The {Expressive} {Power} of {Transformers} with {Chain} of {Thought}},
    url = {http://arxiv.org/abs/2310.07923},
    urldate = {2025-07-25},
    publisher = {arXiv},
    author = {Merrill, William and Sabharwal, Ashish},
    month = apr,
    year = {2024},
}

@article{tversky_judgment_1974,
    title = {Judgment under {Uncertainty}: {Heuristics} and {Biases}: {Biases} in judgments reveal some heuristics of thinking under uncertainty.},
    volume = {185},
    issn = {0036-8075, 1095-9203},
    shorttitle = {Judgment under {Uncertainty}},
    url = {https://www.science.org/doi/10.1126/science.185.4157.1124},
    language = {en},
    number = {4157},
    urldate = {2025-03-27},
    journal = {Science},
    author = {Tversky, Amos and Kahneman, Daniel},
    month = sep,
    year = {1974},
    
    pages = {1124--1131},
}

@incollection{portoraroAutomatedReasoning2025,
  title = {Automated {{Reasoning}}},
  booktitle = {The {{Stanford Encyclopedia}} of {{Philosophy}}},
  author = {Portoraro, Frederic},
  editor = {Zalta, Edward N. and Nodelman, Uri},
  year = {2025},
  edition = {Summer 2025},
  publisher = {Metaphysics Research Lab, Stanford University},
  url = {https://plato.stanford.edu/archives/sum2025/entries/reasoning-automated/},
  urldate = {2025-10-16}
}

@book{murphy_probabilistic_2022,
    address = {Cambridge, Massachusetts London, England},
    series = {Adaptive computation and machine learning},
    title = {Probabilistic machine learning: an introduction},
    isbn = {978-0-262-04682-4 978-0-262-36930-5},
    shorttitle = {Probabilistic machine learning},
    language = {eng},
    publisher = {The MIT Press},
    author = {Murphy, Kevin P.},
    year = {2022},
    
}

@misc{stechly_beyond_2025,
    title = {Beyond {Semantics}: {The} {Unreasonable} {Effectiveness} of {Reasonless} {Intermediate} {Tokens}},
    shorttitle = {Beyond {Semantics}},
    url = {http://arxiv.org/abs/2505.13775},
    urldate = {2025-05-22},
    publisher = {arXiv},
    author = {Stechly, Kaya and Valmeekam, Karthik and Gundawar, Atharva and Palod, Vardhan and Kambhampati, Subbarao},
    month = may,
    year = {2025},
}

@inproceedings{peng_limitations_2024,
    title = {On {Limitations} of the {Transformer} {Architecture}},
    url = {https://par.nsf.gov/servlets/purl/10580944},
    language = {English},
    author = {Peng, Binghui and Narayanan, Srini and Papadimitriou, Christos},
    year = {2024},
    
}

@misc{mondorfAccuracyEvaluating2024,
    title = {Beyond {Accuracy}: {Evaluating} the {Reasoning} {Behavior} of {Large} {Language} {Models} -- {A} {Survey}},
    shorttitle = {Beyond {Accuracy}},
    url = {http://arxiv.org/abs/2404.01869},
    doi = {10.48550/arXiv.2404.01869},
    urldate = {2024-09-09},
    publisher = {arXiv},
    author = {Mondorf, Philipp and Plank, Barbara},
    month = aug,
    year = {2024},
    note = {arXiv:2404.01869 [cs]
Read\_Status: Done
Read\_Status\_Date: 2024-10-25T07:58:55.307Z},
}
